%% file: root.tex
\documentclass[letterpaper, 10 pt, conference]{ieeeconf}  

\IEEEoverridecommandlockouts                              

\usepackage[hidelinks]{hyperref} 

\usepackage{graphicx} 
\usepackage{amsmath} 
\usepackage{amssymb}  
\usepackage{algorithm,algorithmicx} 
\usepackage{algpseudocode}
\usepackage{cleveref} 
\usepackage{xspace} 
\usepackage{xcolor}
\usepackage{subcaption} 
\usepackage{tablefootnote} 
\usepackage{makecell} 

\input{macro.tex}

\crefname{section}{Sec.}{Secs.} 
\crefname{figure}{Fig.}{Figs.} 
\crefname{equation}{}{}
\crefname{algorithm}{Alg.}{Algs.}
\crefname{table}{Tab.}{Tabs.}

\title{\LARGE \bf
Robust and Modular Multi-Limb Synchronization in Motion Stack for Space Robots with Trajectory Clamping via Hypersphere
}

\author{Elian Neppel$^{1}$, Ashutosh Mishra$^{1}$, Shamistan Karimov$^{1}$, \\ Kentaro Uno$^{1}$, Shreya Santra$^{1}$, and Kazuya Yoshida$^{1}$
\thanks{
This work was supported by JST SPRING, Grant Number JPMJSP2114, and JST Moonshot R\&D Program, Grant Number JPMJMS223B.
    }%
\thanks{$^{1}$E. Neppel, A. Mishra, S. Karimov,  K. Uno, S. Santra and K. Yoshida are with the Space Robotics Lab. (SRL) in Department of Aerospace Engineering, Graduate School of Engineering, Tohoku University, Sendai 980--8579, Japan. (E-mail: \tt{neppel.elian.s6@dc.tohoku.ac.jp})  }%
\thanks{
    }
}

\begin{document}

\maketitle
\thispagestyle{empty}
\pagestyle{empty}

\begin{abstract}
	Modular robotics holds immense potential for space exploration, where reliability, repairability, and reusability are critical for cost-effective missions. Coordination between heterogeneous units is paramount for precision tasks -- whether in manipulation, legged locomotion, or multi-robot interaction. Such modular systems introduce challenges far exceeding those in monolithic robot architectures.
	This study presents a robust method for synchronizing the trajectories of multiple heterogeneous actuators, adapting dynamically to system variations with minimal system knowledge.
	This design makes it inherently robot-agnostic, thus highly suited for modularity.
	To ensure smooth trajectory adherence, the multidimensional state is constrained within a hypersphere representing the allowable deviation. The distance metric can be adapted hence, depending on the task and system under control, deformation of the constraint region is possible.
	This approach is compatible with a wide range of robotic platforms and serves as a core interface for \MS, our new open-source universal framework for limb coordination (available at \url{https://github.com/2lian/Motion-Stack}). The method is validated by synchronizing the end-effectors of six highly heterogeneous robotic limbs, evaluating both trajectory adherence and recovery from significant external disturbances.
\end{abstract}

\section{INTRODUCTION}

\todo{
\begin{enumerate}
	\item {check that it fits in 2 pages: OK without this TODO list}
	\item {flush last page using iros template command}
	\item {add synchro citation instead of other cites}
	\item {Video is to be uploaded later}
\end{enumerate}
}

Modular robotics is a rapidly evolving field, with growing potential for space exploration~\cite{ModularReconfig2020,Yim2003ModularRR,Yim2009}. The ability to reconfigure robotic systems dynamically is significant for future extraterrestrial missions, where reliability, repairability, and reusability are critical for reducing costs~\cite{space_busi_2020, space_reconf2021, AHMADZADEH201527}. In these extreme environments, ensuring system adaptability and longevity is paramount. Moreover, robots capable of transforming and adapting to various tasks, unlock numerous possibilities over traditional, monolithic systems~\cite{robotics1986, robot_math_book, ModMan2020, Greel2010, saber2021}.

\begin{figure}[tb]
    \vspace{2mm}
	\centering
	\includegraphics[width=0.75\columnwidth, trim={380 20 520 0}, clip]{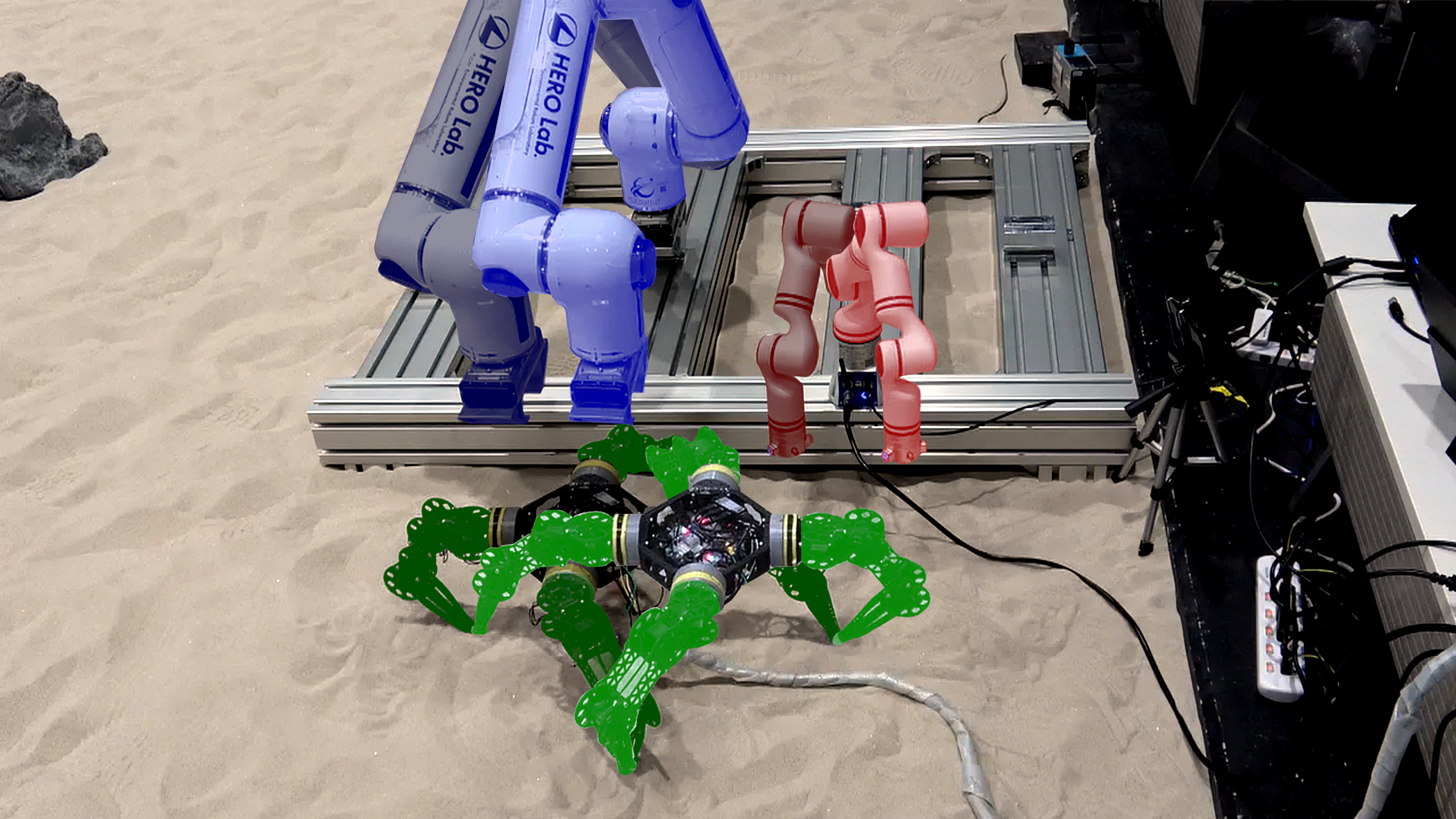}
	\caption{Six heterogeneous limbs Synchronously performing a trajectory. Blue: \MBH; Green: \MBZ; Red: \RM.}
	\label{fig:overview}
\end{figure}

In the context of modular robotics, our goal is the synchronization of multiple heterogeneous actuators shown \cref{fig:overview,tab:robot_limbs}, adapting in real-time to changes in system behavior, without requiring detailed system-specific information~\cite{colab_overview, doi:10.1177/1687814016659597, HAUSER2020103467, distributed_modular_2013, ModularReconfig2020}. This capability is crucial for coordinated locomotion and manipulation with multi-limbed robots, where precise synchronization of \ee and joints is essential for stability and precision~\cite{AdaptiveControl2022, MoveIt2019, robot_math_book, robotics1986}.

\begin{figure}[tb]
    \vspace{2mm}
	\centering
	\begin{subfigure}{0.7\columnwidth}
		\centering
		\includegraphics[width=\columnwidth, trim={0 20 0 160}, clip]{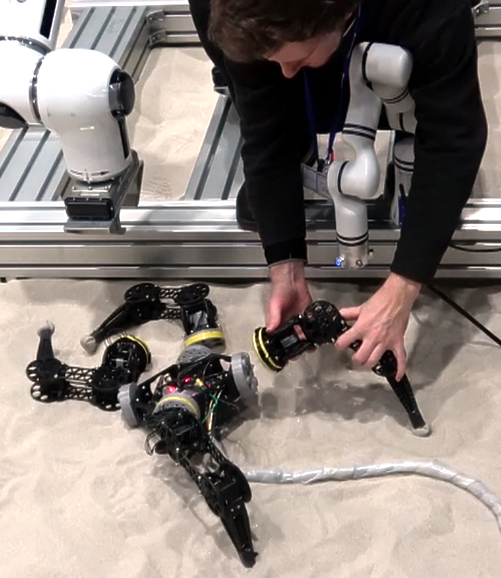}
		\caption{Disassembling limbs.}
		\label{fig:distu2}
		\vspace{0.7em}
	\end{subfigure}
	\hfill
	\begin{subfigure}{0.35\columnwidth}
        \centering
		\includegraphics[width=\columnwidth, trim={0 0 100 150}, clip]{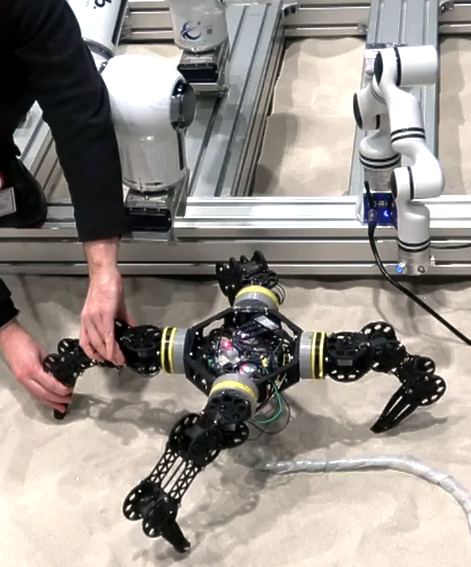}
		\caption{Blocking limbs.}
		\label{fig:distu1}
	\end{subfigure}
	\hfill
	\begin{subfigure}{0.4\columnwidth}
		\centering
		\includegraphics[width=\columnwidth, trim={0 0 30 0}, clip]{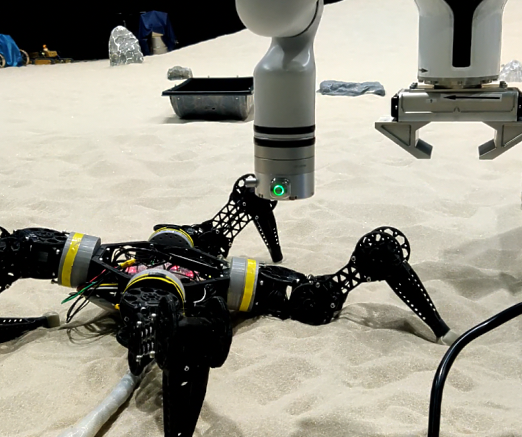}
		\caption{Powering off limbs.}
		\label{fig:distu3}
	\end{subfigure}
	\hfill
	\caption{Trajectory of the limbs \cref{fig:overview}, being purposefully disrupted during the 12-minute robustness experiment.}
	\label{fig:distu}
\end{figure}

Space being our focus, the system must endure harsh conditions and degradation, while remaining safe and operational. Our current and previous experiments, conducted over several months at the JAXA lunar sand field facility~\cite{jaxafield} allow us to work in such difficult environment where hardware quickly degrades. The robustness of our proposed controller is thus critically important. Unlike traditional motion planners, which rely on open-loop execution~\cite{robot_math_book, robotics1986, robotics2008}, our controller incorporates adaptive clamping techniques for real-time trajectory correction and recovery. This ensures that the system safely adheres to trajectories despite significant disturbances, such as actuator disconnections or defects illustrated \cref{fig:distu}.

\begin{table}[tb]
    \vspace{2mm} 
	\caption{Characteristics of the robotic limbs.}
	\label{tab:robot_limbs}
	\centering
	\renewcommand{\arraystretch}{1.2} 
	\begin{tabular}{lcccccc}
		\hline
		\textbf{Limbs} & \textbf{Joints} & \textbf{Speed} & \textbf{Mass} & \textbf{Computer} \\
		\hline
		\MBH           & 7               & 0.15 rad/s     & 20 kg         & Onboard           \\
		\MBZ ×4        & 3               & 10 rad/s       & 300 g         & Main PC           \\
		\RM            & 7               & 3.1 rad/s      & 9 kg          & External          \\
		\hline
	\end{tabular}
\end{table}

Building on the flexibility and distributed structure of \rt~\cite{ros2}, our new open-source \MS framework is inspired by the robot-agnostic approach of \moit~\cite{MoveIt2019}. This paper's proposed controller serves as the high-level application programming interface (API) for the \MS. Our software’s robot-agnostic nature allows it to control a wide variety of robotic platforms, facilitating rapid prototyping and ensuring clear separation of concerns~\cite{programming2007,ros2}. Through extensive experimentation, we have validated the controller’s effectiveness in synchronizing six heterogeneous limbs \cref{tab:robot_limbs,fig:overview}, showcasing its robust, robot-agnostic capabilities. Through our previously developed modular \mBZ~\cite{moobot_zero} we could induce major disturbances and study the response to both nominal and degraded conditions \cref{fig:distu}.

\section{METHOD}

Requirements for our controller are as follows, the output command should:
\begin{enumerate}
	\item Never be far from the sensed state, for safety reason.
	\item Be part of the trajectory.
	\item Advance in the trajectory.
\end{enumerate}


\subsection{System-Agnostic Mathematical Definition}

\label{sec:math}

Let $Y_k \in M$ denote the current system state at the step k, $U_k \in M$ the command, $S \in M$ the starting position of the trajectory, $F \in M$ the final position of the trajectory.
For requirement 2, the trajectory from $S$ to $F$ is given as \cref{eq:traj} by a function $\mathcal{T}$.

\begin{equation}
	\begin{split}
		\mathcal{T} : [0,1]\times M\times M \to {M}, \\
		\quad \mathcal{T}(0, S, F) = S, \quad \mathcal{T}(1, S, F) =  F
	\end{split} \label{eq:traj}
\end{equation}

To define requirement 1, we use a metric space $\mathcal{M}=(M, d)$ which, by definition, is the association of a set $M$ with a distance $d: (M, M) \to \mathbb{R}$~\cite{robot_math_book}. We leverage the arbitrary distance function $d$, to clamp onto a hypersphere $V_k$ of center $Y_k$ defined \cref{eq:Vk}. This allows us to change the shape of the hypersphere in accordance with the system and task at hand.

If using $M=\mathbb{R}^3$, a sensible choice is to define the trajectory as the segment $[S, F]$, and the metric as Euclidean norm $d=\ell^2$ with components weighted by a factor $\Delta_e \in \mathbb{R}^3$. Given \cref{eq:eucl} with $\div$ as the element-wise division, such metric results in $\mathcal{V}_k$ being a stretched 3D sphere.

\begin{gather}
	\mathcal{V}_k = \{X \in \mathcal{M} \mid {d}(X, Y_k) \leq 1 \}\label{eq:Vk}\\
	{d}_{eucl}(X_1, X_2)_k = \|(X_1 - X_2) \div \Delta_e\|_2 \label{eq:eucl}
\end{gather}

Our previous requirements can now be rigorously formulated, the output command
\begin{enumerate}
	\item Belongs to the hypersphere $\mathcal{V}_k$.
	\item Belongs to $\{\mathcal{T}(t, S, F) \mid t \in [0, 1]\}$
	\item Maximizes $t$.
\end{enumerate}

These conditions ensure that the command $U_k$ moves as close as possible to $F$, while respecting the trajectory and staying close to the current state $Y_k$. The clamped command can be derived as \cref{eq:tmax,eq:clamping}. This clamping method preserves the trajectory’s path $\mathcal{T}$, while potentially altering its execution speed, hence adapting to the system’s performance.

\begin{gather}
	t_k^{max} = \arg\max_{t \in [0, 1]}\{ t \mid \mathcal{T}(t, S, F) \in \mathcal{V}_k\} \label{eq:tmax} \\
	U_k = \mathcal{T}( t_k^{max}, S, F) \label{eq:clamping}
\end{gather}

\subsection{Algorithm and Edge Cases}
\label{sec:methalgo}

The algorithmic procedure is detailed \cref{alg:clamped_ik}. Depending on the choice of $\mathcal{T}$ and $d$, the \textit{for loop} line~3 is vectorizable, leading to fast computation~\cite{cuda, algo_book}.

\begin{algorithm}[ht]
	\caption{General Hypersphere Clamping}
	\label{alg:clamped_ik}
	\begin{algorithmic}[1]
		\State $I$: number of trajectory samples


		\Procedure{HyperSphereClamp}{$Y_k$, $S$, $F$ $\mathcal{T}$, d, $I$}
		\For{$t = 1 \to 0 \quad \textbf{in  } I \textbf{  steps}$}
		\State $p =\mathcal{T}(t, S, F) $
		\State $dist =d(p, Y_k) $
		\If{$dist \leq 1$}
		\State \Return $p$
		\EndIf
		\EndFor
		\State \Return $None$
		\EndProcedure
	\end{algorithmic}
\end{algorithm}

One unaddressed edge case is the behavior in case no solution exists -- return value is $None$. This can happen during external disturbances shown \cref{fig:distu}, bringing the state $Y_k$ far from the trajectory. In that case, a recovery strategy should be executed at a higher level. We propose a non-exhaustive list of strategies

\begin{itemize}
	\item Start a trajectory toward the last valid $p$ that was returned, then continue the previous trajectory. (Strategy chosen for our implementation)
	\item Return the $p$ with the smallest $dist$ value.
	\item Start a new trajectory from $Y_k$ to $F$.
\end{itemize}

A final rare edge case is the fact that nothing prevents the trajectory from jumping back in time $t$. This can happen when the system is forcefully driven backward, similarly to \cref{fig:distu1}. Depending on the application, one might need to ensure that $t$ never decreases between results of \cref{alg:clamped_ik}.

\subsection{Single End-Effector Implementation}

\label{sec:single}

This section will use $M=SE(3)$, the Special 3-dimensional Euclidean group~\cite{robot_math_book,choice_metrics}. With $g \in SE(3)$ represented using a translation $\mathbf{v} \in \mathbb{R}^3$ and rotation $\mathbf{R} \in SO(3)$, noted as $g=(\mathbf{v}_g, \quad \mathbf{R}_g)$.

Our trajectory is an interpolation from $S$ to $F$ given as \cref{eq:trajsingle}. The vector part uses Linear interpolation (LERP) and the rotational part uses Spherical Linear interpolation (SLERP)~\cite{robot_math_book,awsome_quaterion_article}.

\begin{align}
	\mathcal{T}_{SE}(t, S, F) =
	(\quad
	 & (1 - t) \mathbf{v}_S + t \mathbf{v}_F ,\nonumber \\
	 & \text{SLERP}(\mathbf{R}_S, \mathbf{R}_F, t)
	\quad) \label{eq:trajsingle}
\end{align}

Metrics $d$ for $SE(3)$ are numerous with different properties depending on the application, furthermore at least two scaling factors for the translational part and rotational part are necessary~\cite{robot_math_book,choice_metrics}.

Given $p_e \in \mathbb{R}$ the allowed translational error and $r_e \in \mathbb{R}$ the allowed rotational error, \cref{eq:met1} defines one distance metric suited for our application: the Euclidean norm of, the vector divided by $p_e$, augmented with the angle of the rotation $\theta(\mathbf{R})$ divided by $r_e$. Note that, this metric is general and could be refined to better fit certain tasks.

\begin{equation}
	d_{SE}(g_a, g_b) =  \left\|
	\begin{bmatrix}
		(\mathbf{v}_b - \mathbf{v}_a)/p_e \\
		\theta(\mathbf{R}_a^{-1} \cdot \mathbf{R}_b)/r_e
	\end{bmatrix} \right\|_2 \\
	\label{eq:met1}
\end{equation}

Once $U_k$ computed using \cref{eq:clamping,alg:clamped_ik}, the target pose is sent to the inverse kinematics (IK) controller of the associated limb.

\subsection{Multiple \EEs Implementation}
\label{sec:meth_multi}

Our proposition for multiple \ees consists in vertically stacking the single \ee implementation of \cref{sec:single}. Using $n$ as the number of end-effectors, we define $M = SE(3)^n$, the trajectory as \cref{eq:trajsen} and metric as \cref{eq:metsen}.

\begin{gather}
	\mathcal{T}(t, S, F) =
	\begin{bmatrix}
		\mathcal{T}_{SE}(t, S_1, F_1) \\
		\mathcal{T}_{SE}(t, S_2, F_2) \\
		...                           \\
		\mathcal{T}_{SE}(t, S_n, F_n)
	\end{bmatrix} \label{eq:trajsen}
\end{gather}

The metric \cref{eq:met1} is used to map $SE(3)^n$ onto the Euclidean n-space, then a k-norm $\ell^k$ can be applied, thus giving the distance $d : SE(3)^n \times SE(3)^n \to \mathbb{R}^n \to \mathbb{R}$ detailed in \cref{eq:metsen}. For our application, we chose $\ell^{\infty}$.

\begin{gather}
	d(X, Y) =
	\left\|
	\begin{bmatrix}
		d_{SE}(X_1, Y_1) \\
		d_{SE}(X_2, Y_2) \\
		...              \\
		d_{SE}(X_n, Y_n)
	\end{bmatrix}
	\right\|_{\infty}
	\label{eq:metsen}
\end{gather}

Lastly, \cref{alg:clamped_ik} requires a discretization parameter $I$, representing the sample count. For LERP trajectories, we propose \cref{eq:stepsize} to deduce a suitable $I$ based on the distance between $S$ and $F$ under the chosen metric $d$. \textit{step\_distance} represents the spacing between consecutive samples along the trajectory. We recommend using 0.01, which results in approximately 100 samples per hypersphere radius.

\begin{equation}
	I = d(S,F) / \text{step\_distance} \label{eq:stepsize}
\end{equation}

\section{EXPERIMENTS AND RESULTS}
\label{sec:res}

\subsection{\EE Driven Outside Its Motion Range}
\label{sec:meth_single}

\begin{figure}[b]
	\centering
	\begin{subfigure}{0.24\columnwidth}
		\centering
		\includegraphics[width=\columnwidth, trim={15 15 15 15}, clip]{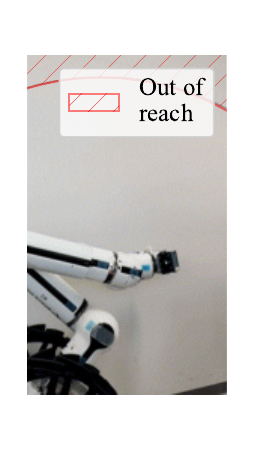}
		\caption{Start.}
		\label{fig:ik_real01}
	\end{subfigure}
	\begin{subfigure}{0.24\columnwidth}
		\centering
		\includegraphics[width=\columnwidth, trim={15 15 15 15}, clip]{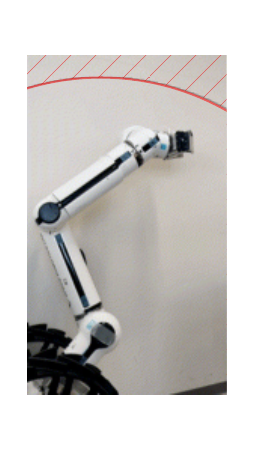}
		\caption{Raising.}
		\label{fig:ik_real02}
	\end{subfigure}
	\begin{subfigure}{0.24\columnwidth}
		\centering
		\includegraphics[width=\columnwidth, trim={15 15 15 15}, clip]{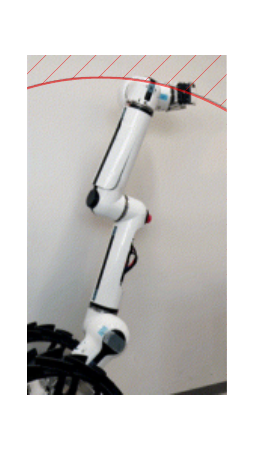}
		\caption{Out of reach.}
		\label{fig:ik_real03}
	\end{subfigure}
	\begin{subfigure}{0.24\columnwidth}
		\centering
		\includegraphics[width=\columnwidth, trim={15 15 15 15}, clip]{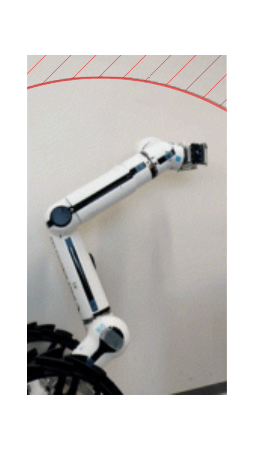}
		\caption{Lowering.}
		\label{fig:ik_real04}
	\end{subfigure}
	\caption{Snapshots of the \mBH safely pausing a trajectory partially outside its range of motion.}
	\label{fig:ikreal}
\end{figure}

\begin{figure}[tb]
    \vspace{2mm}
	\centering
	\includegraphics[width=\columnwidth]{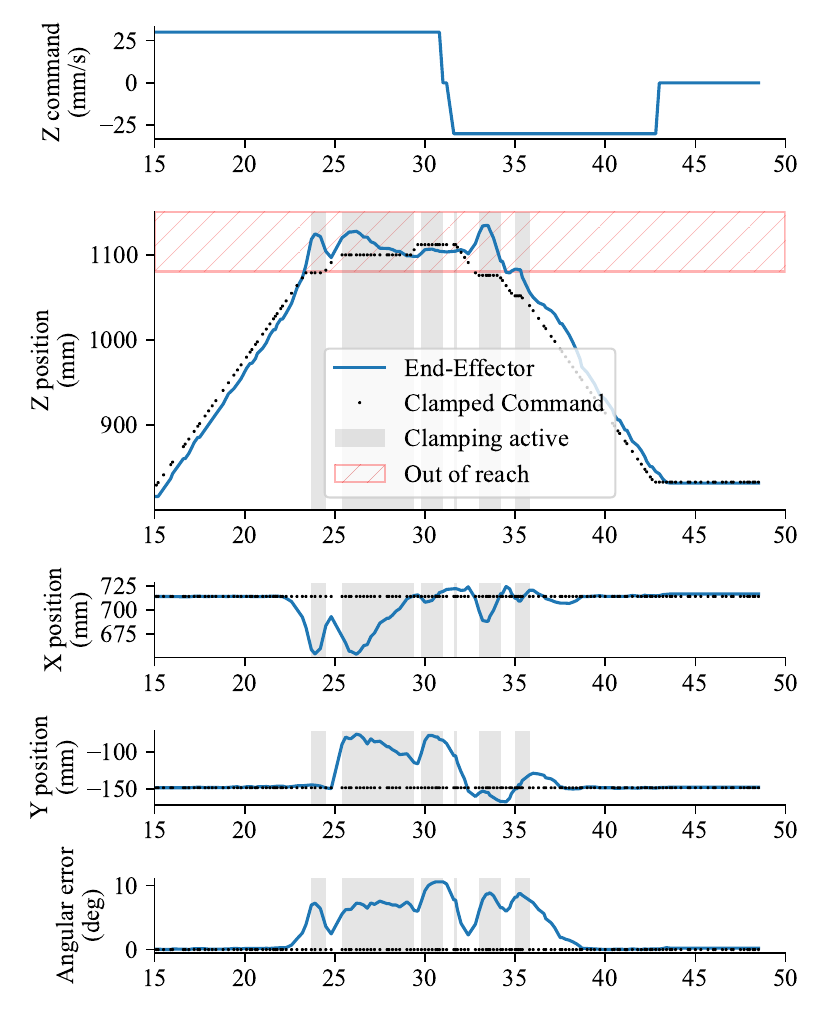}
	\caption{\MBH driven outside its range of motion, by a pure $z$ speed trajectory. Clamping pauses the impossible movement, thus preserving the trajectory's path.}
	\label{fig:ik_op}
\end{figure}

Our first experiment assesses the robustness of the single end-effector application of \cref{sec:single,sec:meth_multi}. One \mBH is driven at a constant speed upward, outside the kinematic range of the arm, then it is driven back down in range. The goal is to assess the robustness and recovery capabilities when faced with an abnormal system response. The results can be seen \cref{fig:ik_op,fig:ikreal}.

Using $Y_k \in M = SE(3)^1$, $\mathcal{T}$ and $d$ defined \cref{eq:trajsen,eq:metsen}, $\dot{P}$ as the input speed, $t_k$ as the time, \cref{eq:contspeed} defines the controller used for this section. The clamping is recursively performed while integrating the speed into a position. The parameters of \cref{eq:metsen} were set at $p_e=50 \text{ mm}$ and $r_e=30 \text{ deg}$. 

With the goal of testing solely the robustness of the clamping, contrary to the recommendations of \cref{sec:methalgo}, no recovery strategies were implemented. Therefore, if no solution exists, the controller waits.

\begin{multline}
	U_{k+1} =
	\mathit{HyperSphereClamp}(Y_k, U_k, \\
	U_k \cdot (t_{k+1} - t_k) \dot{P}, d, I )
	\label{eq:contspeed}
\end{multline}

Before clamping \cref{fig:ik_real01,fig:ik_real02,fig:ik_op}, the constant speed trajectory from the clamped command is excelent. The motion is a straight line going up at 30 mm/s.

Around 22 s on \cref{fig:ik_op}, the end-effector begins diverging from the unreachable command. Despite the input to go up, the controller rightfully clamps $z$. \cref{fig:ik_real03,fig:ik_op} show the motion slowing and stopping in order to adhere to the trajectory and not worsen the situation.

Later around 31 s, \cref{fig:ik_real04,fig:ik_op}, the speed input goes back down toward the mobility area of the \mBH. Some clamping continues due to the IK slowly moving away from a singularity~\cite{robot_math_book}. Soon, the the trajectory recovers, going back down through the exact same path it went up.

\subsection{Multiple Heterogeneous \EEs}

\label{sec:res_mult}

\begin{figure}[tb]
    \vspace{2mm}
	\centering
	\begin{subfigure}{0.49\columnwidth}
		\centering
		\includegraphics[width=\columnwidth]{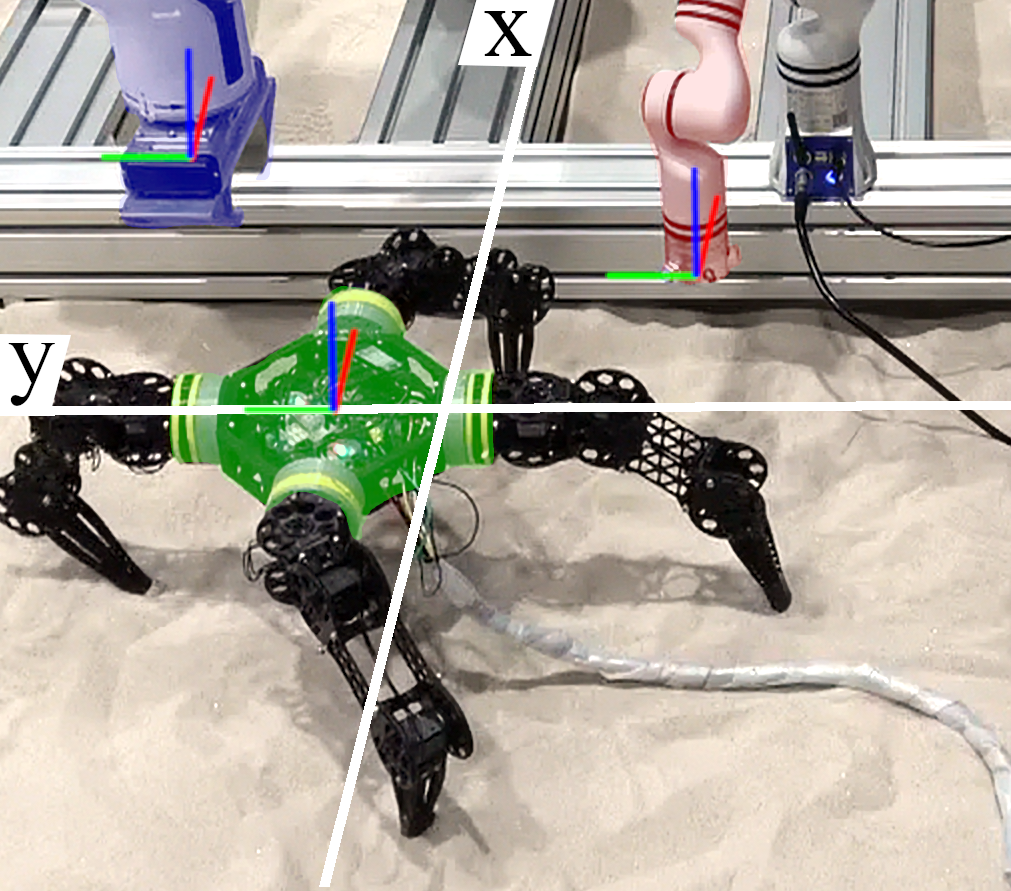}
		\caption{Left corner.}
		\label{fig:sq1}
	\end{subfigure}%
	\hfill
	\begin{subfigure}{0.49\columnwidth}
		\centering
		\includegraphics[width=\columnwidth]{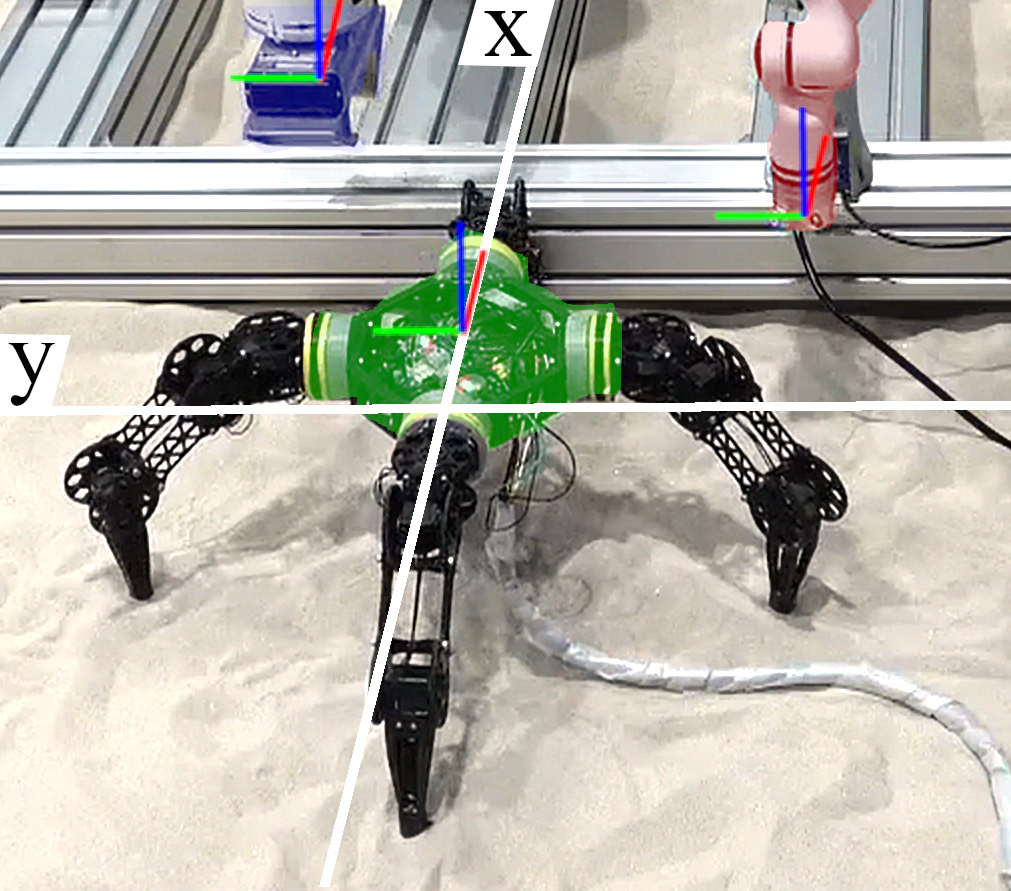}
		\caption{Front corner.}
		\label{fig:sq4}
	\end{subfigure}

	\vspace{0.2cm}

	\begin{subfigure}{0.49\columnwidth}
		\centering
		\includegraphics[width=\columnwidth]{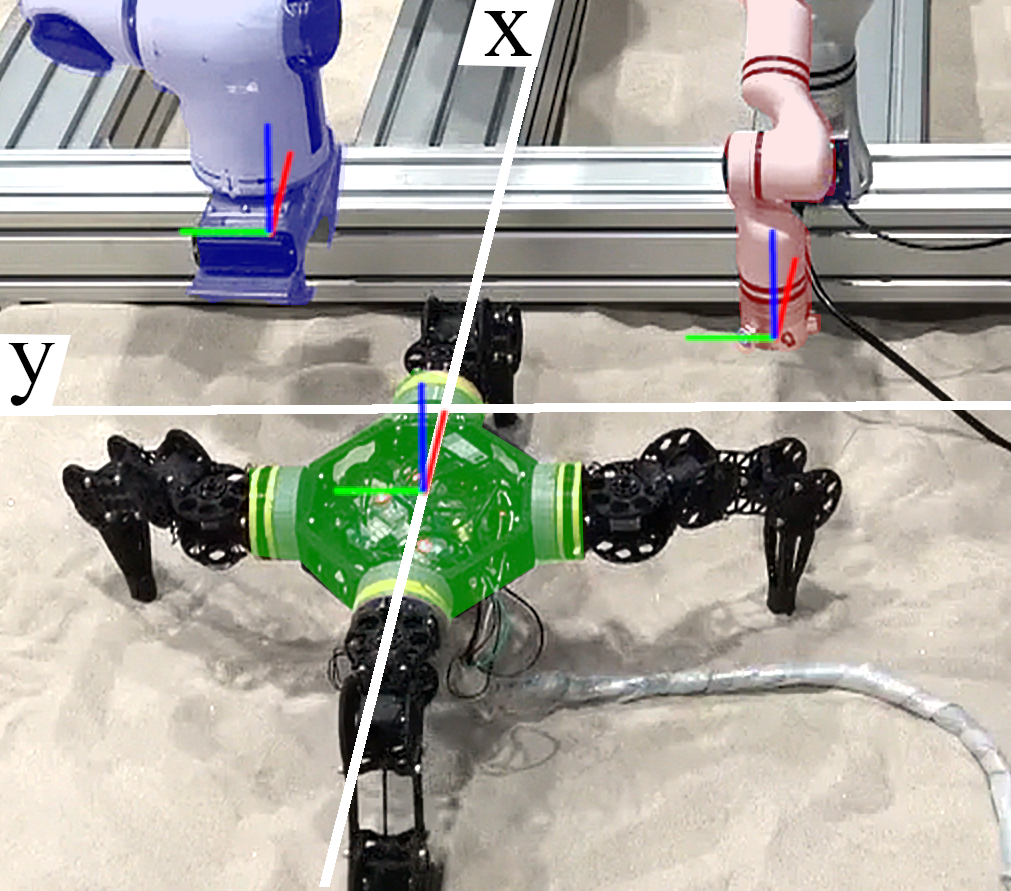}
		\caption{Back corner.}
		\label{fig:sq3}
	\end{subfigure}%
	\hfill
	\begin{subfigure}{0.49\columnwidth}
		\centering
		\includegraphics[width=\columnwidth]{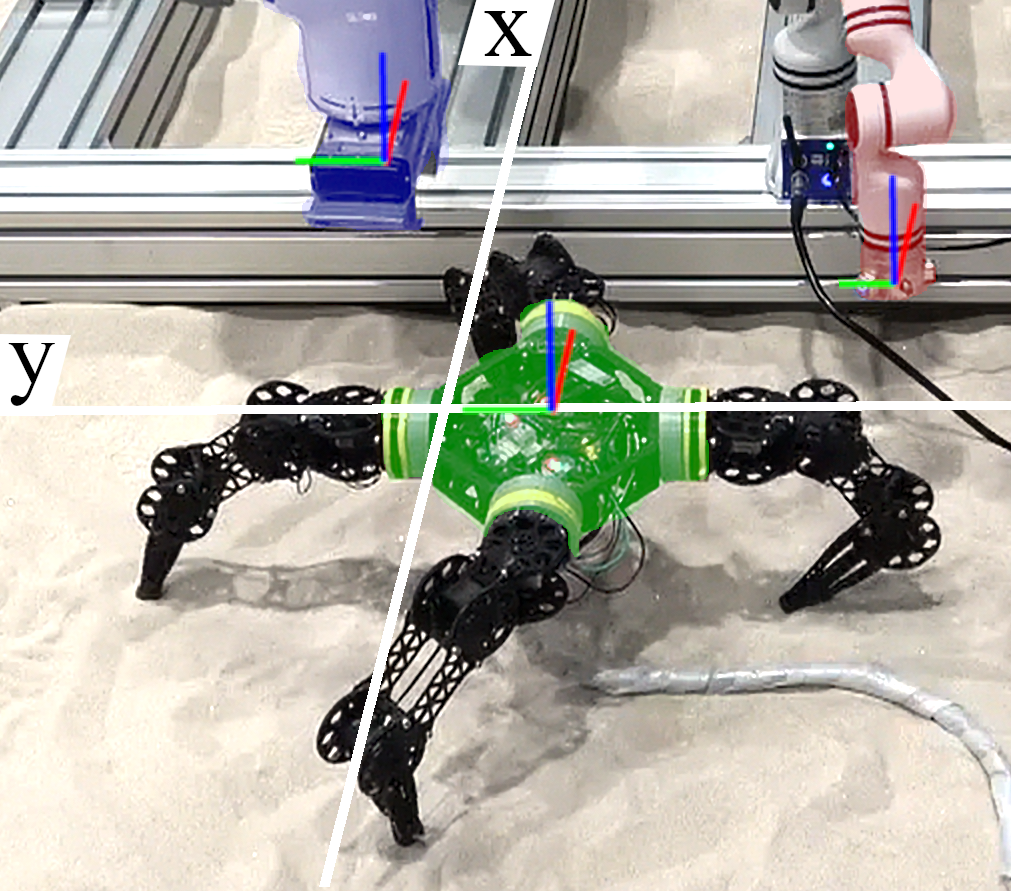}
		\caption{Right corner.}
		\label{fig:sq2}
	\end{subfigure}
	\caption{Four corners of the square trajectory, synchronously performed by \MBH (Blue), \MBZ (Green), and \RM (Red).
	}
	\label{fig:sq_main}
\end{figure}

To test our multi \ees Implementation \cref{sec:meth_multi}, we used the diverse limbs detailed \cref{tab:robot_limbs} and shown \cref{fig:overview,fig:sq_main}. Those present widely different characteristics. Furthermore, computers -- handling IK commands and forward kinematics (FK) states -- are distributed over the network, further inducing latency and instability.
The \mBZ \ees lacking degrees of freedom to control their rotation, the parameters of \cref{eq:metsen} were set at $r_e=+\infty$ and a precise $p_e=20 \text{ mm}$.

The trajectory of each end-effector $\mathcal{T}_{SE}$ is linearly interpolated between 4 waypoints using \cref{eq:trajsen,eq:trajsingle}. Those waypoints form a square of diagonal 200 mm. \cref{fig:sq1,fig:sq2,fig:sq3,fig:sq4} capture all limbs at each four corners of the trajectory, while \cref{fig:traj_xy} later plots it for one limb.

For the experimental setup, the \mBZ is horizontally rotated by $180$ deg. This results in the \mBZ main body, \mBH and \rM moving together along $x, y$ relative to the ground. \cref{fig:sq_main} shows this synchronicity.

\renewcommand\theadalign{bc}
\renewcommand\theadfont{\bfseries}

\begin{table}[tb]
    \vspace{2mm}
	\centering
	\caption{Provoked disruptions during the unsupervised 12-minute robustness experiment.}
	\label{tab:robustness_experiment}
	\renewcommand{\arraystretch}{1.2} 
	\begin{tabular}{lccc}
		\hline
        \textbf{Disruption Type}    & \thead{Robot} & \thead{Limb\\count} & \thead{Occurrences} \\
		\hline
		Mechanical blockage         & \MBZ           & 1                   & 8                    \\
		Mechanical slowdown         & \MBZ           & 1                   & 2                    \\
		Disassembly                 & \MBZ           & 1 to 3              & 10                   \\
		Powering off (causing fall) & \MBZ           & 4                   & 7                    \\
		Powering off                & \RM            & 1                   & 1                    \\
		IK computation error        & \MBH           & 1                   & 1                    \\
		\hline
	\end{tabular}
\end{table}

To assess robustness, the trajectory was set on repeat for 12 minutes, and all software was left untouched with no human input. Numerous mechanical and power disruptions were provoked as per \cref{tab:robustness_experiment} and illustrated \cref{fig:distu1,fig:distu2,fig:distu3}.

Despite these numerous disturbances the controller always recovered, safely continuing the trajectory without any human supervision. The experiment was terminated after 12 minutes, although it could have continued further.

\subsection{Analysis of a Recovery from Power Loss}
\label{sec:res_disr}

\begin{figure}[tb]
    \vspace{2mm}
	\centering
	\begin{subfigure}{0.7\columnwidth}
		\centering
		\includegraphics[width=\columnwidth]{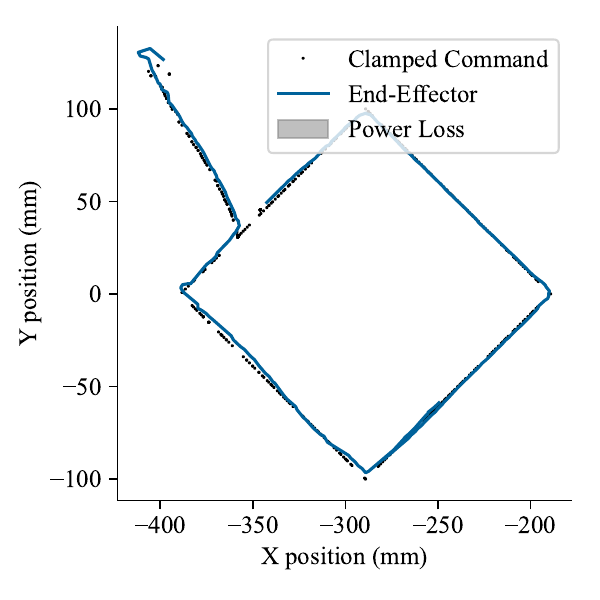}
		\caption{\MBZ leg \#3. Time independent plot between 710-730 s.}
		\label{fig:traj_xy}
	\end{subfigure}
	\vspace{1em}
	\begin{subfigure}{0.99\columnwidth}
		\centering
		\includegraphics[width=\columnwidth]{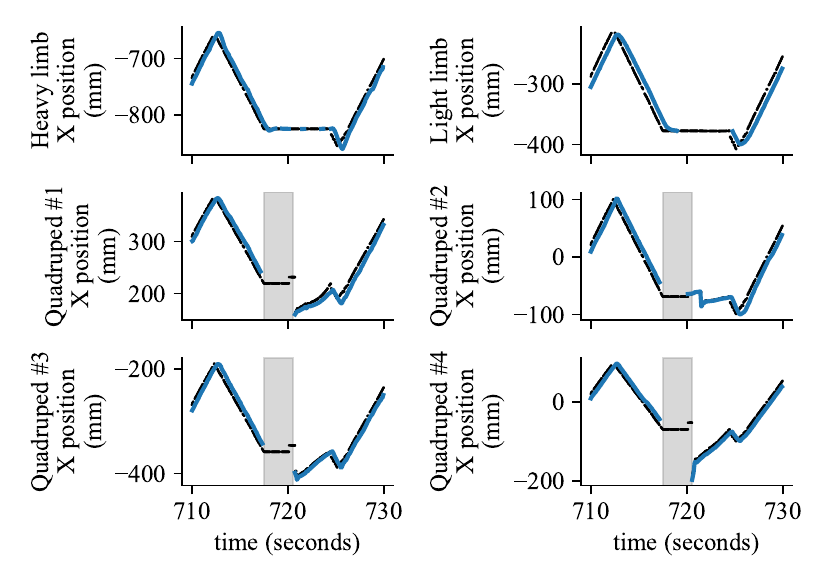}
		\caption{All limb positions through time.}
		\label{fig:traj_zt}
	\end{subfigure}
	\caption{Position and command during square trajectory execution, with 4 limbs experiencing and recovering from power loss. \Ee data line is discontinuous when power is lost.}
	\label{fig:traj_main}
\end{figure}

During the experiment (\cref{sec:res_mult}), between the 718th and 720th seconds, we cut power to four limbs of the \mBZ. This causes the robot to fall before standing back up and continuing the trajectory \cref{fig:distu3,fig:traj_main}.

Such disturbance causes a spike in the \ee position of all four affected limbs (\cref{fig:traj_xy}). While powered off, the state $Y_k$ no longer updates, as visible in \cref{fig:traj_zt}. However, the command still moves slightly, with a maximum deviation of $p_e = 20$ mm. When power is restored at 720 seconds, $Y_k$ abruptly refreshes to its post-fall sensor positions. The clamped command quickly corrects before smoothly returning onto the planned trajectory -- consistent with the recovery strategy outlined in \cref{sec:methalgo}.

This recovery process occurs simultaneously across all affected limbs. As shown in \cref{fig:traj_zt}, all four disturbed limbs recover in a manner similar to \cref{fig:traj_xy}, effectively causing the robot to stand back up. Meanwhile, the two unaffected limbs safely pause their motion, avoiding further disturbances. Once all limbs recover onto the trajectory around the 725th second, they seamlessly resume the synchronized motion.

\section{DISCUSSION}

\subsection{System Agnosticism and Modularity}

Our controller demonstrates strong robot-agnostic behavior. As described in \cref{sec:math}, the only system-specific requirement is the metric function $d$. \cref{sec:meth_multi} simplifies this down to only a simple pair of numbers, yet able to fully control an arbitrary number of end-effectors.

The synchronization of the 6 \ees in \cref{sec:res_mult} was remarkably seamless, despite the significant differences between the robots outlined in \cref{tab:robot_limbs}. Moreover, the system’s recovery capabilities in \cref{sec:res_disr} are truly impressive. Typically, stopping the modules while the \mBZ is down, then making it stand back up before resuming the trajectory, would require an intricate state machine and numerous health checks. However, our approach bypasses this complexity, greatly simplifying the system and reducing failure points.

These properties, combined with a robust mathematical foundation, make our controller an ideal fit for the API of our open-source, robot-agnostic \MS. Any set of robotic limbs can be moved through identical API calls, enabling fast development and strong separation of concerns~\cite{MoveIt2019, ros2, programming2007}. Beyond this paper, multi-joint synchronization has also been implemented as API using our controller, and more than six of our different robot modules are being controlled through the \MS.

This system-agnostic approach is crucial for modular robotics and, in our experience, greatly benefits collaboration, research, and development.

\subsection{Robustness}

Our controller is designed for extreme environments, such as extraterrestrial robotic operations. In these settings, safety is prioritized over performances, demanding resilience to significant system degradation~\cite{space_busi_2020,space_reconf2021,moobot_zero,modeling2021}.

Experiments confirm the controller's robustness and safety (\cref{tab:robustness_experiment,sec:res}). It sustained over 12 minutes of disruptions across six limbs, including disconnections, physical blockages, and sensor errors. As shown in \cref{sec:res_mult,sec:res_disr}, recovery remains smooth and controlled, even under severe disturbances. Furthermore, our robots have been using the controller as their API through several month-long field-tests in the JAXA lunar facility~\cite{jaxafield} without fail.

These capabilities are not solely due to the proposed controller, but also stem from the \MS framework, which smoothly controls joints and IK. This paper represents one core component of this software.

Unlike traditional, performance focused motion systems, that rely on planning before open-loop execution ~\cite{MoveIt2019,robotics2008}, our approach incorporates adaptive clamping, allowing real-time trajectory rectification. Our application \cref{sec:meth_single}, closes the loop, only when required, i.e. when clamping is applied. This effectively creates a highly flexible semi-open-loop. Existing solutions typically require learning-based techniques to achieve such adaptivity~\cite{ModularReconfig2020,robotics2008,moobot_zero}, which demands expensive training and simulations. In contrast, our method provides deterministic, real-time adaptation and recovery without reliance on black-box processes.

\section{CONCLUSIONS}

\addtolength{\textheight}{-0.0cm}   

This work presents a robust and modular multi-actuator synchronization method for space robotics that uses a hypersphere-based trajectory clamping mechanism to ensure reliable execution. The proposed method is inherently robot-agnostic, requiring only a simple distance metric as system-specific input, making it ideal for flexible software and development.

Key contributions of this work include a novel adaptive clamping strategy that keeps the multi-dimensional state within safe bounds at all times, a system-agnostic architecture enabling on the fly coordination and recovery of heterogeneous limb modules, and extensive experimental validation on six vastly different robotic limbs. This level of flexibility marks a significant step forward in ensuring reliability and safety in modular space missions.

Looking ahead, several promising research directions emerge. One avenue is the integration of noisy sensors and actuators with uncertainty models to adapt the trajectory based on real-time precision and external feedback. Additionally, we are exploring the synchronization of modular force-controlled modules for compliant manipulation and body motion. Future work will also focus on deploying the method across a broader range of coordinated systems and improving trajectory definition for smoother motion.

Finally, as Motion Stack matures and releases as open source framework, we will continue to develop and share its components for diverse applications in modular robotics. By pursuing these directions, we seek to enhance the flexibility and robustness of multi-limb robotic systems. Ensuring they can support long-duration, high stake, and reconfigurable missions in orbit or on planetary surfaces in the years to come.


\bibliography{bib.bib}

\end{document}

%% file: macro.tex
\newcommand{\MS}{Motion~Stack\xspace}

\newcommand{\mBH}{heavy limb\xspace}
\newcommand{\mBZ}{quadruped\xspace}
\newcommand{\rM}{light limb\xspace}
\newcommand{\MBH}{Heavy limb\xspace}
\newcommand{\MBZ}{Quadruped\xspace}
\newcommand{\RM}{Light limb\xspace}
\newcommand{\moit}{\textit{Moveit!}\xspace}
\newcommand{\rt}{ROS2\xspace}

\newcommand{\ee}{end-effector\xspace}
\newcommand{\ees}{end-effectors\xspace}
\newcommand{\Ee}{End-effector\xspace}

\newcommand{\EE}{End-Effector\xspace}
\newcommand{\EEs}{End-Effectors\xspace}

\newcommand\todo[1]{} 

%% file: root.bbl
\begin{thebibliography}{10}
\providecommand{\url}[1]{#1}
\csname url@samestyle\endcsname
\providecommand{\newblock}{\relax}
\providecommand{\bibinfo}[2]{#2}
\providecommand{\BIBentrySTDinterwordspacing}{\spaceskip=0pt\relax}
\providecommand{\BIBentryALTinterwordstretchfactor}{4}
\providecommand{\BIBentryALTinterwordspacing}{\spaceskip=\fontdimen2\font plus
\BIBentryALTinterwordstretchfactor\fontdimen3\font minus
  \fontdimen4\font\relax}
\providecommand{\BIBforeignlanguage}[2]{{%
\expandafter\ifx\csname l@#1\endcsname\relax
\typeout{** WARNING: IEEEtran.bst: No hyphenation pattern has been}%
\typeout{** loaded for the language `#1'. Using the pattern for}%
\typeout{** the default language instead.}%
\else
\language=\csname l@#1\endcsname
\fi
#2}}
\providecommand{\BIBdecl}{\relax}
\BIBdecl

\bibitem{ModularReconfig2020}
\BIBentryALTinterwordspacing
G.~Liang, D.~Wu, Y.~Tu, and T.~L. Lam, ``Decoding modular reconfigurable
  robots: A survey on mechanisms and design,'' 2023. [Online]. Available:
  \url{https://arxiv.org/abs/2310.09743}
\BIBentrySTDinterwordspacing

\bibitem{Yim2003ModularRR}
\BIBentryALTinterwordspacing
M.~H. Yim, K.~Roufas, D.~Duff, Y.~Zhang, C.~Eldershaw, and S.~B. Homans,
  ``Modular reconfigurable robots in space applications,'' \emph{Autonomous
  Robots}, vol.~14, pp. 225--237, 2003. [Online]. Available:
  \url{https://api.semanticscholar.org/CorpusID:9796019}
\BIBentrySTDinterwordspacing

\bibitem{Yim2009}
\BIBentryALTinterwordspacing
M.~Yim, P.~White, M.~Park, and J.~Sastra, \emph{Modular Self-Reconfigurable
  Robots}.\hskip 1em plus 0.5em minus 0.4em\relax New York, NY: Springer New
  York, 2009, pp. 5618--5631. [Online]. Available:
  \url{https://doi.org/10.1007/978-0-387-30440-3_334}
\BIBentrySTDinterwordspacing

\bibitem{space_busi_2020}
\BIBentryALTinterwordspacing
A.~Orlova, R.~Nogueira, and P.~Chimenti, ``The present and future of the space
  sector: A business ecosystem approach,'' \emph{Space Policy}, vol.~52, p.
  101374, 2020. [Online]. Available:
  \url{https://www.sciencedirect.com/science/article/pii/S0265964620300163}
\BIBentrySTDinterwordspacing

\bibitem{space_reconf2021}
A.~M. Romanov, V.~D. Yashunskiy, and W.-Y. Chiu, ``A modular reconfigurable
  robot for future autonomous extraterrestrial missions,'' \emph{IEEE Access},
  vol.~9, pp. 147\,809--147\,827, 2021.

\bibitem{AHMADZADEH201527}
\BIBentryALTinterwordspacing
H.~Ahmadzadeh and E.~Masehian, ``Modular robotic systems: Methods and
  algorithms for abstraction, planning, control, and synchronization,''
  \emph{Artificial Intelligence}, vol. 223, pp. 27--64, 2015. [Online].
  Available:
  \url{https://www.sciencedirect.com/science/article/pii/S0004370215000260}
\BIBentrySTDinterwordspacing

\bibitem{robotics1986}
M.~H. Raibert and E.~R. Tello, ``Legged robots that balance,'' \emph{IEEE
  Expert}, vol.~1, no.~4, pp. 89--89, 1986.

\bibitem{robot_math_book}
R.~M. Murray, S.~S. Sastry, and L.~Zexiang, \emph{A Mathematical Introduction
  to Robotic Manipulation}, 1st~ed.\hskip 1em plus 0.5em minus 0.4em\relax USA:
  CRC Press, Inc., 1994.

\bibitem{ModMan2020}
A.~Yun, D.~Moon, J.~Ha, S.~Kang, and W.~Lee, ``Modman: An advanced
  reconfigurable manipulator system with genderless connector and automatic
  kinematic modeling algorithm,'' \emph{IEEE Robotics and Automation Letters},
  vol.~5, no.~3, pp. 4225--4232, July 2020.

\bibitem{Greel2010}
K.~Tadakuma, R.~Tadakuma, A.~Maruyama, E.~Rohmer, K.~Nagatani, K.~Yoshida,
  A.~Ming, M.~Shimojo, M.~Higashimori, and M.~Kaneko, ``Mechanical design of
  the wheel-leg hybrid mobile robot to realize a large wheel diameter,'' in
  \emph{2010 IEEE/RSJ International Conference on Intelligent Robots and
  Systems}, Oct 2010, pp. 3358--3365.

\bibitem{saber2021}
A.~M. Romanov, V.~D. Yashunskiy, and W.-Y. Chiu, ``Saber: Modular
  reconfigurable robot for industrial applications,'' in \emph{2021 IEEE 17th
  International Conference on Automation Science and Engineering (CASE)}, Aug
  2021, pp. 53--59.

\bibitem{colab_overview}
Z.~Feng, G.~Hu, Y.~Sun, and J.~Soon, ``An overview of collaborative robotic
  manipulation in multi-robot systems,'' \emph{Annual Reviews in Control},
  vol.~49, pp. 113--127, 2020.

\bibitem{doi:10.1177/1687814016659597}
\BIBentryALTinterwordspacing
J.~Liu, X.~Zhang, and G.~Hao, ``Survey on research and development of
  reconfigurable modular robots,'' \emph{Advances in Mechanical Engineering},
  vol.~8, no.~8, p. 1687814016659597, 2016. [Online]. Available:
  \url{https://doi.org/10.1177/1687814016659597}
\BIBentrySTDinterwordspacing

\bibitem{HAUSER2020103467}
\BIBentryALTinterwordspacing
S.~Hauser, M.~Mutlu, P.-A. Léziart, H.~Khodr, A.~Bernardino, and A.~Ijspeert,
  ``Roombots extended: Challenges in the next generation of self-reconfigurable
  modular robots and their application in adaptive and assistive furniture,''
  \emph{Robotics and Autonomous Systems}, vol. 127, p. 103467, 2020. [Online].
  Available:
  \url{https://www.sciencedirect.com/science/article/pii/S0921889019303379}
\BIBentrySTDinterwordspacing

\bibitem{distributed_modular_2013}
\BIBentryALTinterwordspacing
D.~J. Christensen, U.~P. Schultz, and K.~Stoy, ``A distributed and
  morphology-independent strategy for adaptive locomotion in
  self-reconfigurable modular robots,'' \emph{Robotics and Autonomous Systems},
  vol.~61, no.~9, pp. 1021--1035, 2013. [Online]. Available:
  \url{https://www.sciencedirect.com/science/article/pii/S0921889013001061}
\BIBentrySTDinterwordspacing

\bibitem{AdaptiveControl2022}
\BIBentryALTinterwordspacing
S.~Aoi, P.~Manoonpong, Y.~Ambe, F.~Matsuno, and F.~Wörgötter, ``Adaptive
  control strategies for interlimb coordination in legged robots: A review,''
  \emph{Frontiers in Neurorobotics}, vol.~11, 2017. [Online]. Available:
  \url{https://www.frontiersin.org/journals/neurorobotics/articles/10.3389/fnbot.2017.00039}
\BIBentrySTDinterwordspacing

\bibitem{MoveIt2019}
M.~Görner, R.~Haschke, H.~Ritter, and J.~Zhang, ``Moveit! task constructor for
  task-level motion planning,'' in \emph{2019 International Conference on
  Robotics and Automation (ICRA)}, 2019, pp. 190--196.

\bibitem{jaxafield}
{JAXA}, ``{JAXA}’s {R}\&{D} facility for open innovation and space
  exploration projects,'' \url{ https://www.ihub-tansa.jaxa.jp/english/ },
  2017, (Accessed: February 16, 2025).

\bibitem{robotics2008}
\BIBentryALTinterwordspacing
W.~Chung, L.-C. Fu, and S.-H. Hsu, \emph{Motion Control}.\hskip 1em plus 0.5em
  minus 0.4em\relax Berlin, Heidelberg: Springer Berlin Heidelberg, 2008, pp.
  133--159. [Online]. Available:
  \url{https://doi.org/10.1007/978-3-540-30301-5_7}
\BIBentrySTDinterwordspacing

\bibitem{ros2}
S.~Macenski, T.~Foote, B.~Gerkey, C.~Lalancette, and W.~Woodall, ``Robot
  operating system 2: Design, architecture, and uses in the wild,''
  \emph{Science Robotics}, vol.~7, no.~66, p. 6074, 2022.

\bibitem{programming2007}
\BIBentryALTinterwordspacing
P.~A. Laplante, \emph{What Every Engineer Should Know About Software
  Engineering}.\hskip 1em plus 0.5em minus 0.4em\relax Boca Raton: Taylor \&
  Francis, 2007, includes bibliographical references and index. [Online].
  Available: \url{https://www.worldcat.org/oclc/1285845379}
\BIBentrySTDinterwordspacing

\bibitem{moobot_zero}
D.~Ai, T.~Sinsunthorn, P.~Pama, S.~Santra, K.~Uno, and K.~Yoshida, ``{Moonbot
  0: Design and Development of a Modular Robot for Lunar Exploration and
  Assembly Tasks},'' in \emph{The 68th Society of Systems, Control and
  Information Engineers Research Presentation Conference}, May 2024.

\bibitem{cuda}
D.~B. Kirk and W.-m.~W. Hwu, \emph{Programming Massively Parallel Processors: A
  Hands-on Approach}, 3rd~ed.\hskip 1em plus 0.5em minus 0.4em\relax San
  Francisco, CA, USA: Morgan Kaufmann Publishers Inc., 2012.

\bibitem{algo_book}
T.~H. Cormen, C.~E. Leiserson, R.~L. Rivest, and C.~Stein, \emph{Introduction
  to Algorithms, Third Edition}, 3rd~ed.\hskip 1em plus 0.5em minus 0.4em\relax
  The MIT Press, 2009.

\bibitem{choice_metrics}
\BIBentryALTinterwordspacing
M.~Žefran, V.~Kumar, and C.~Croke, ``{Choice of Riemannian Metrics for Rigid
  Body Kinematics},'' in \emph{International Design Engineering Technical
  Conferences and Computers and Information in Engineering Conference}, vol.
  Volume 2B: 24th Biennial Mechanisms Conference, 08 1996, p. V02BT02A030.
  [Online]. Available: \url{https://doi.org/10.1115/96-DETC/MECH-1148}
\BIBentrySTDinterwordspacing

\bibitem{awsome_quaterion_article}
\BIBentryALTinterwordspacing
K.~Shoemake, ``Animating rotation with quaternion curves,'' \emph{SIGGRAPH
  Comput. Graph.}, vol.~19, no.~3, p. 245–254, July 1985. [Online].
  Available: \url{https://doi.org/10.1145/325165.325242}
\BIBentrySTDinterwordspacing

\bibitem{modeling2021}
\BIBentryALTinterwordspacing
J.~Thangavelautham and Y.~Xu, \emph{Modeling Excavation, Site Preparation, and
  Construction of a Lunar Mining Base Using Robot Swarms}.\hskip 1em plus 0.5em
  minus 0.4em\relax ASCE Library, 2021, pp. 1310--1325. [Online]. Available:
  \url{https://ascelibrary.org/doi/abs/10.1061/9780784483374.121}
\BIBentrySTDinterwordspacing

\end{thebibliography}
